\documentclass[sigconf]{acmart}

\AtBeginDocument{%
  }
\usepackage{booktabs}
\usepackage{graphicx}
\usepackage{siunitx}
\usepackage{algorithm}
\usepackage{algpseudocode}
\usepackage{dblfloatfix}
\usepackage{makecell}
\usepackage[skins,breakable]{tcolorbox}

\copyrightyear{2025}
\acmYear{2025}
\setcopyright{cc}
\setcctype{by-nc-sa}
\acmConference[CIKM '25]{Proceedings of the 34th ACM International
Conference on Information and Knowledge Management}{November 10--14,
2025}{Seoul, Republic of Korea}
\acmBooktitle{Proceedings of the 34th ACM International Conference on
Information and Knowledge Management (CIKM '25), November 10--14, 2025,
Seoul, Republic of Korea}\acmDOI{10.1145/3746252.3760848}
\acmISBN{979-8-4007-2040-6/2025/11}
\author{Yujin Park}
\affiliation{%
  \orcid{https://orcid.org/0009-0001-8988-5698}
  \institution{Hanyang University}
  \city{Seoul}
  \postcode{04763}
  \country{Republic of Korea}
}
\email{yujin1019a@hanyang.ac.kr}

\author{Haejun Chung\textsuperscript{*}}
\affiliation{%
\orcid{https://orcid.org/0000-0001-8959-237X}
  \institution{Hanyang University}
  \city{Seoul}
  \postcode{04763}
  \country{Republic of Korea}
}
\email{haejun@hanyang.ac.kr}

\author{Ikbeom Jang\textsuperscript{*}}
\affiliation{%
\orcid{https://orcid.org/0000-0002-6901-983X}
  \institution{Hankuk University of Foreign Studies}
  \city{Yongin}
  \postcode{17035}
  \country{Republic of Korea}
}
\email{ijang@hufs.ac.kr}

\renewcommand{\shortauthors}{Park et al.}

\thanks{\textsuperscript{*}Corresponding authors.}

\begin{document}

\title{EZ-Sort: Efficient Pairwise Comparison via Zero-Shot CLIP-Based Pre-Ordering and Human-in-the-Loop Sorting}

\begin{abstract}
Pairwise comparison is often favored over absolute rating or ordinal classification in subjective or difficult annotation tasks due to its improved reliability; however, exhaustive comparisons require a massive number of annotations ($O(n^2)$). Recent work ~\cite{jang2022decreasing} greatly reduced the annotation burden ($O(n\log n)$) by actively sampling pairwise comparisons using a sorting algorithm. 
We further improve annotation efficiency by 1) roughly pre-ordering items using the Contrastive Language-Image Pre-training (CLIP) model hierarchically without training and 2) replacing easy, obvious human comparisons with automated comparisons.
The proposed EZ-Sort first produces a CLIP‑based zero‑shot \emph{pre‑ordering}, then initializes bucket-aware Elo scores, and finally runs an uncertainty‑guided human-in-the-loop MergeSort.
Validation was conducted using various datasets: face‑age estimation (FGNET)~\cite{993553}, historical image chronology (DHCI)~\cite{palermo2012dating}, and retinal image quality assessment (EyePACS)~\cite{fu2019evaluation}. It showed that EZ‑Sort reduced human annotation cost by 90.5\% compared to exhaustive pairwise comparisons and by 19.8\% compared to prior work~\cite{jang2022decreasing} (when $n=100$) while improving or maintaining inter-rater reliability.
These results demonstrate that combining CLIP-based priors with uncertainty-aware sampling yields an efficient and scalable solution for pairwise ranking. 
Code available at \url{https://github.com/yujinPark02/EZ-Sort-CIKM2025}.

\end{abstract}

\ccsdesc[300]{Information systems~Data labeling}
\ccsdesc[300]{Computing methodologies~Ranking}

\keywords{Pairwise comparison, Human-in-the-loop sorting, VLM-based pre-ordering, Annotation, Labeling}

\begin{teaserfigure}
  \centering
  \includegraphics[width=\textwidth, keepaspectratio]{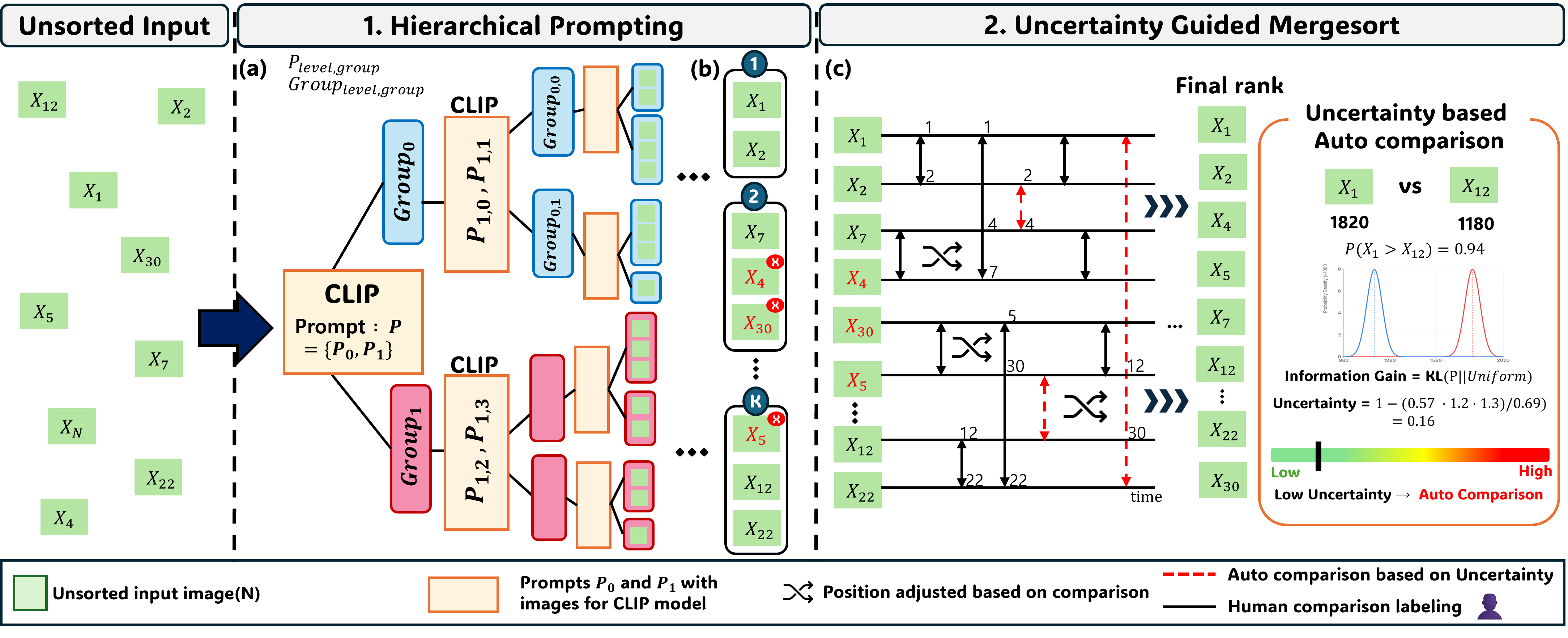}
  \caption{\textbf{Overview of the EZ-Sort framework.}
  The framework operates in three stages:  
  (\textbf{a}) CLIP-based zero-shot \emph{pre-ordering} is performed via hierarchical prompting that recursively groups unsorted images using binary prompts.  
  (\textbf{b}) The resulting fine-grained groups are merged into $k$ coarse buckets, and each image is assigned an Elo score based on its bucket ID and CLIP confidence.  
  (\textbf{c}) An uncertainty-aware MergeSort selectively routes high-uncertainty comparisons to human annotators while automatically resolving confident ones.
  }
  \label{fig:teaser}
\end{teaserfigure}

\maketitle

\section{Introduction}
Pairwise comparison is widely preferred in subjective annotation tasks, including perceptual quality assessment, face-age estimation, and medical image triage~\cite{kalpathy2016plus}, due to its superior inter-rater reliability compared to absolute or ordinal ratings. However, exhaustive pairwise labeling incurs a quadratic annotation burden $O(n^2)$, quickly becoming infeasible as dataset sizes grow.
This annotation bottleneck has tangible implications for large-scale clinical diagnostics, population health studies, and the preservation of historical records, where subject-matter expertise is scarce and annotation budgets are limited. This scalability challenge can be addressed by leveraging computational sorting algorithms, such as MergeSort.

However, traditional methods, such as the Bradley–Terry–Luce model~\cite{bradley1952rank} and active learning strategies~\cite{maystre2017just,saar2004active}, typically assume uniform priors and overlook opportunities to leverage the existing semantic structure in the data. More recently, sorting-based approaches have incorporated active query selection to reduce the number of comparisons~\cite{jang2022decreasing}. 

To further alleviate the annotation burden, we propose incorporating vision-language models (VLM), such as CLIP~\cite{radford2021learning}, to provide a strong starting point for the sorting process. Because such models are pre-trained on hundreds of millions of image-text pairs, proper prompts enable a coarse initial ranking of given items to be labeled. This leads to a significant reduction in the number of comparisons needed for sorting. We iteratively execute this process hierarchically to improve accuracy. We also propose replacing human comparisons with automated comparisons for item pairs with low uncertainty. By combining this prior knowledge with an uncertainty-guided comparison selection strategy, we aim to provide a complementary perspective to existing methods, focusing on the efficient utilization of both model priors and human expertise.

Our key contributions include (1) leveraging pre-trained vision-language priors to reduce the initial annotation search space significantly, (2) applying this VLM-based pre-ordering hierarchically for improved accuracy, and (3) introducing a novel uncertainty-guided sorting strategy to prioritize human annotation resources intelligently.

\begin{figure}
    \centering
    \includegraphics[width=\linewidth, height=0.22\textheight]{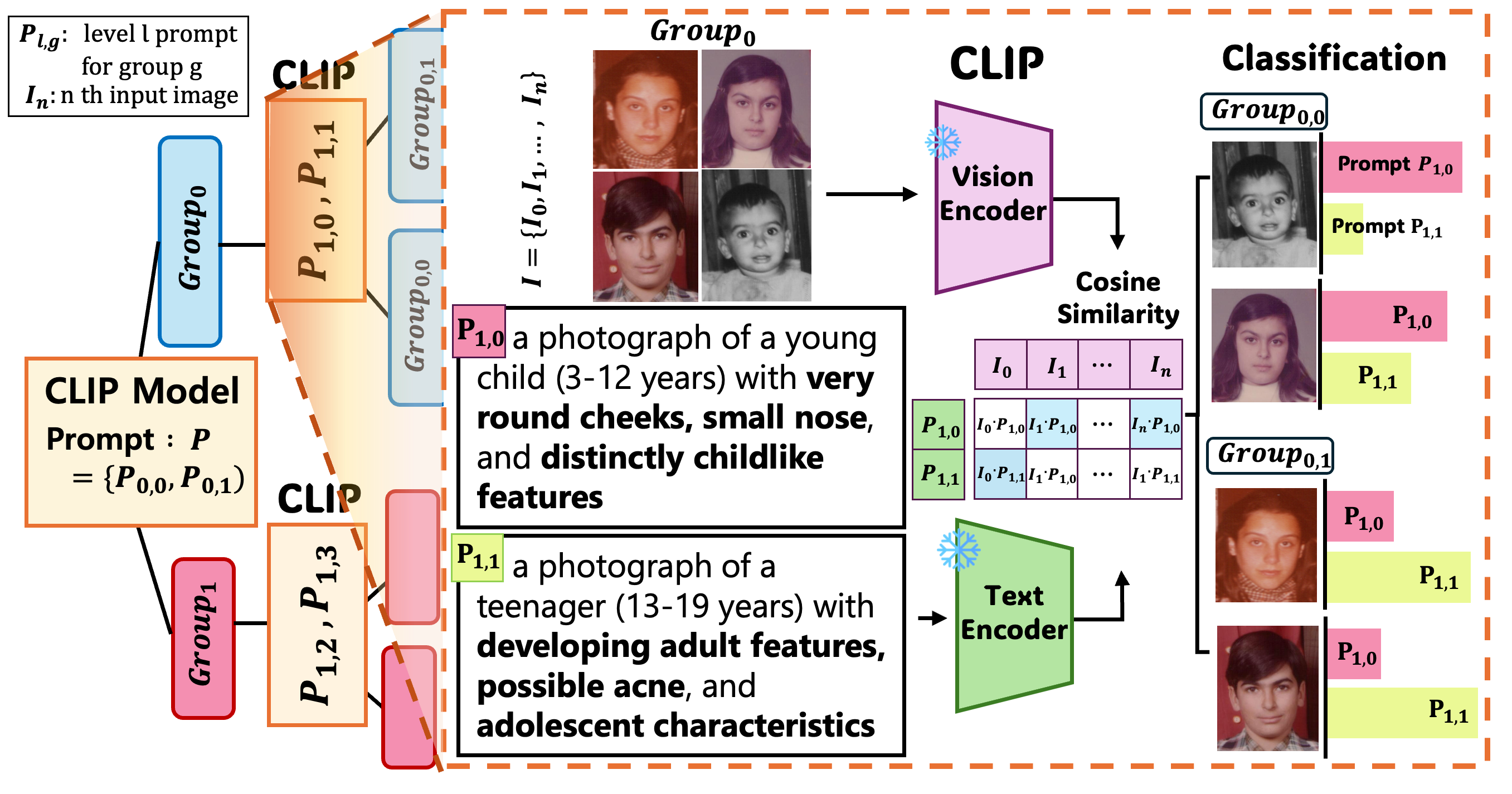}
     \caption{\textbf{CLIP-based hierarchical classification at Level~1.}
Input images are compared against two age-related prompts (child vs. teenager) using CLIP cosine similarity and assigned to subgroups based on the highest score.}
    \vspace{-12pt}
    \label{fig:clip_level1}
\end{figure}

\section{Methods}
\label{sec:method}
EZ-Sort follows a three-stage pipeline: (i) hierarchical CLIP prompting provides a zero-shot semantic pre-ordering (i.e., ordinal classification) of images, (ii) bucket-aware Elo scores initialize priors, and (iii) Kullback-Leibler(KL) based MergeSort routes only uncertain pairs to annotators.~\cite{budagam2024hierarchical,booch2021thinking}.

This pipeline reflects a dual-process structure, with automatic, low-uncertainty comparisons (System 1) and human deliberation for high-uncertainty cases (System 2). The subsequent subsections provide a detailed overview of each stage.

\subsection{Hierarchical Prompt Design for Zero-Shot Classification}

Traditional classification often struggles with ambiguous subjective attributes. To address this, we introduce a hierarchical prompting strategy that exploits CLIP's zero-shot capabilities through multi-level binary decisions. Inspired by how machine learning has optimized sorting algorithms~\cite{mankowitz2023faster,zhao2018n,bai2023sorting}, our method enhances classification robustness by decomposing complex decisions into more tractable binary ones.

To evaluate its benefit over single-level prompting, we tested two flat baselines: a minimal template 
$P_{\text{simple}} = \{ \text{``a photograph of a } c \text{''} \}$ and an enhanced variant incorporating GPT-4-generated attributes 
$P_{\text{flat-detailed}} = \{ \text{``a photograph of } c \text{ with } \mathcal{A}_c \text{''} \}$. Experiments across three datasets demonstrate that hierarchical prompts outperform both baselines, yielding an improvement of up to 2.0 MAE. This performance gain arises from (1) \textbf{decision decomposition}, which replaces $k$-way classification with $\log_2 k$ binary steps where CLIP excels, and (2) \textbf{progressive refinement}, which focuses each decision on discriminative features at that granularity. By mimicking human coarse-to-fine reasoning, our hierarchical strategy enhances both interpretability and classification accuracy, particularly in visually ambiguous domains.

\subsubsection{Prompt Generation Methodology}

We replace flat $k$-class classification with an adaptive hierarchical
binary structure, dynamically yielding groups until no meaningful
visual distinctions remain.  At each level~$\ell$, we define
$P_\ell=\{p_{\ell,0},p_{\ell,1},\dots,p_{\ell,2^\ell-1}\}$, building upon recent advances in prompt learning~\cite{zhou2022learning,zhou2022conditional}.

\begin{tcolorbox}[colback=gray!8,colframe=gray!50,
                  left=1.2ex,right=1.2ex,top=0.8ex,bottom=0.8ex]
\textit{\textbf{Automated Hierarchical Prompt Generation}: \\ ''Given the domain \textbf{[domain]} with range \textbf{[range]},  
iteratively divide each group into two visually distinguishable sub-groups using observable anatomical or textural features.
Continue dividing until no further meaningful visual distinctions are identifiable.  \textbf{Avoid} behavioural or contextual clues.''}
\end{tcolorbox}

For face-age estimation (\emph{domain = face},
\emph{range = 0–60+ years}) this produces prompts such as
\textit{``rounded cheeks, large forehead''} (infants) versus
\textit{``defined cheekbones, mature jawline''} (adults); the recursion
stops once visual distinctions become ambiguous.

\textbf{Adaptive Depth Criterion.}
Depth is not fixed; generation halts when GPT-4 signals that further splits are unreliable, resulting in 3 to 5 levels in practice, balancing granularity with annotation load.

\textbf{Group Assignment.}
Binary outcomes $c_{i,\ell}\!\in\!\{0,1\}$ are combined into a final group index
\begin{equation}
  g_i=\sum_{\ell=1}^{d_i} c_{i,\ell}\,2^{\ell-1},
\end{equation}
where $d_i$ is the image-specific depth. This hierarchical encoding stabilizes Elo initialization and reduces the need for downstream human comparisons.

\subsubsection{CLIP-Based Classification}

For each level in the hierarchy, we compute text and image embeddings using CLIP~\cite{radford2021learning}, and measure similarity between image $i$ and prompt $j$ as cosine similarity: $s_{i,l,j} = \frac{\mathbf{v}_i \cdot \mathbf{t}_{l,j}}{||\mathbf{v}_i|| \cdot ||\mathbf{t}_{l,j}||}$.
This approach leverages CLIP's zero-shot classification capabilities~\cite{qian2024online} and builds upon knowledge-enhanced visual models~\cite{shen2022k}.

Classification decisions and confidence scores are derived from these similarities: $c_{i,l} = \arg\max_j s_{i,l,j}$ and $\text{conf}_{i,l} = \frac{\exp(s_{i,l,c_{i,l}}/\tau)}{\sum_j \exp(s_{i,l,j}/\tau)}$ where $\tau = 0.1$ is a temperature parameter controlling the softness of the distribution. Figure~\ref{fig:clip_level1} shows an example of this process at Level~1 using binary prompts for age grouping. Having established the pre-ordering and initial bucket assignments, we next describe how pairs are selected for human annotation based on uncertainty.

\subsubsection{Bucket-Aware Rating Initialization}

To obtain a coarse ordering for Elo initialization, the fine-grained groups produced by hierarchical classification are merged into \(k\) primary buckets using the mapping
\(M\colon \{0,1,\dots,2^{d}-1\}\!\to\!\{0,1,\dots,k-1\}\) defined by
\(M(g) = \lfloor g\,k / 2^{d} \rfloor\).
This rule uniformly distributes groups while preserving their ordinal relationships. 
We empirically find the optimal number of primary buckets to be \(K\) between 3 and 5, which balances ranking accuracy and annotation cost; the smaller \(k\) merges overly dissimilar items, while larger \(k\) increases unertain cross-bucket comparisons without accuracy benefits.

Accordingly, we set \(k=5\) for CLIP-friendly domains such as FGNET but \(k=3\) for more challenging domains like DHCI and EyePACS, thereby containing noise and comparison overhead.
Each image \(i\) then receives an Elo rating
\[
r_i = r_{\text{base}}(b_i) + \eta_i\bigl(1.5 - \text{conf}_i\bigr),
\]
where \(b_i=M(g_i)\) is its bucket, \(\eta_i\sim\mathcal{U}(-\delta_b,\delta_b)\) adds controlled randomness, and the confidence term keeps high-confidence samples stable while allowing low-confidence ones to move more freely.

\subsubsection{Information-Gain-Based Uncertainty and Comparison Prioritization}

We assess the informativeness of comparisons using the KL-divergence from a uniform prior. For items $i$ and $j$ with Elo scores $r_i$ and $r_j$, the pre-comparison distribution is $P_{\text{before}} = [p_{ij}, 1-p_{ij}]$, where $p_{ij} = \frac{1}{1+10^{(r_j-r_i)/400}}$, following the default setting. 

The information gain is computed as:
\begin{equation}
    \operatorname{InfoGain}(i,j) = \operatorname{KL}(P_{\text{before}} \| P_{\text{uniform}}) = \sum_k p_k \log \frac{p_k}{0.5}
    \label{eq:infogain}
\end{equation}

To prioritize comparisons, we define:
\begin{equation}
    \operatorname{Priority}(i,j) = \operatorname{InfoGain}(i,j) \cdot \gamma(b_i, b_j) \cdot \phi(\text{conf}_i, \text{conf}_j)
    \label{eq:priority}
\end{equation}
where $\gamma(b_i, b_j) = 1.2$ for cross-bucket pairs (to account for CLIP uncertainty), and $1.0$ otherwise; $\phi(\cdot)$ penalizes low-confidence predictions: $\phi(\text{conf}_i, \text{conf}_j) = 2.0 - \text{avg\_conf}$.The uncertainty measure is its complement, normalized by the maximum binary InfoGain ($\log 2$): $ \operatorname{uncertainty}(i,j) = 1 - \frac{\operatorname{Priority}(i,j)}{\log 2} $.

\subsubsection{MergeSort with Uncertainty-Aware Comparison Selection}
Our algorithm follows the \textit{exact} comparison schedule of classical MergeSort; the only difference is how each comparison is resolved.  
Let $\operatorname{uncertainty}(i,j)$ be the KL-based score for items $(i,j)$ and $\theta_t$ an adaptive threshold (Sec.~\ref{sec:theta}).  
During each merge, we apply the rule  
\begin{equation}
  \text{query\_human}(i,j)\;\Longleftrightarrow\;
  \operatorname{uncertainty}(i,j)\;\ge\;\theta_t,                \label{eq:decision}
\end{equation}
breaking ties in favor of the human query.  
If the condition is false, the outcome is decided automatically by
$\operatorname{sign}(r_i-r_j)$, where $r$ denotes the current Elo scores.

Because (i) every pair that standard MergeSort examines is still compared, and  
(ii) deciding \eqref{eq:decision} takes constant time ($O(1)$), the overall complexity remains
$O(n\log n)$. Thus, EZ-Sort preserves the algorithmic optimality while redirecting human effort only to the most uncertain comparisons.

\paragraph{Adaptive threshold}
\label{sec:theta}  
The threshold is adapted based on the remaining budget and current accuracy:
\begin{equation}
    \theta_t = \theta_0
               \Bigl(1+\alpha\,\frac{\text{remaining}}{\text{total}}\Bigr)
               \beta^{\operatorname{accuracy}_t},
    \label{eq:theta}
\end{equation}
where $t$ denotes the current evaluation cycle (0-based), updated after every batch of human comparisons or every $k$ merge operation, whichever occurs first. Here, $\alpha$ controls budget sensitivity, and $0<\beta<1$ encourages increased automation as accuracy improves.

\begin{table*}[t]
\caption{Inter-rater reliability comparison across datasets. Reported metrics: \textit{Sp} (Spearman), \textit{Ke} (Kendall), \textit{Pe} (Pearson), and \textit{ICC} (intraclass correlation). Variance estimates are reported in parentheses where available.}
\small
\setlength{\tabcolsep}{4pt}
\centering
\begin{tabular}{lcccccccccccc}
\toprule
& \multicolumn{4}{c}{\textbf{Retina (EyePACS)}} 
& \multicolumn{4}{c}{\textbf{Historical (DHCI)}} 
& \multicolumn{4}{c}{\textbf{Face (FGNET)}} \\
\cmidrule(lr){2-5} \cmidrule(lr){6-9} \cmidrule(lr){10-13}
\textbf{Method} & Sp & Ke & Pe & ICC & Sp & Ke & Pe & ICC & Sp & Ke & Pe & ICC \\
\midrule
Classification &
0.53 & 0.46 & 0.54 & 0.75 &
0.39 & 0.33 & 0.42 & 0.68 &
0.92 & 0.85 & 0.94 & 0.97 \\[-3pt]
& \scriptsize(±0.06) & \scriptsize(±0.07) & \scriptsize(±0.07) & \scriptsize(N/A) &
\scriptsize(±0.06) & \scriptsize(±0.05) & \scriptsize(±0.06) & \scriptsize(N/A) &
\scriptsize(±0.04) & \scriptsize(±0.09) & \scriptsize(±0.03) & \scriptsize(N/A) \\
Sort comparison~\cite{jang2022decreasing} &
0.72 & 0.56 & 0.72 & 0.89 &
0.47 & 0.35 & 0.47 & \textbf{0.78} &
\textbf{0.97} & 0.88 & \textbf{0.97} & \textbf{0.99} \\[-3pt]
& \scriptsize(±0.07) & \scriptsize(±0.06) & \scriptsize(±0.07) & \scriptsize(N/A) &
\scriptsize(±0.17) & \scriptsize(±0.15) & \scriptsize(±0.17) & \scriptsize(N/A) &
\scriptsize(±0.01) & \scriptsize(±0.01) & \scriptsize(±0.01) & \scriptsize(N/A) \\
EZ-Sort (CIKM) &
\textbf{0.85} & \textbf{0.76} & \textbf{0.85} & \textbf{0.94} &
\textbf{0.47} & \textbf{0.39} & \textbf{0.47} & 0.73 &
0.96 & \textbf{0.91} & 0.96 & 0.99 \\[-3pt]
& \scriptsize(±0.09) & \scriptsize(±0.14) & \scriptsize(±0.09) & \scriptsize(N/A) &
\scriptsize(±0.17) & \scriptsize(±0.15) & \scriptsize(±0.16) & \scriptsize(N/A) &
\scriptsize(±0.01) & \scriptsize(±0.02) & \scriptsize(±0.01) & \scriptsize(N/A) \\
\bottomrule
\end{tabular}
\label{tab:inter-rater}
\end{table*}

\begin{table}[t]
\caption{The number of human annotations (comparisons) required. FGNET dataset is used.}
\centering\small
\resizebox{\linewidth}{!}{%
\begin{tabular}{lccccc}
\toprule
\textbf{Dataset size} &
\makecell[c]{\textbf{Exhaustive} \\ \textbf{comparison}~\cite{thurstone1927method}} &
\makecell[c]{\textbf{Sort} \\ \textbf{comparison}~\cite{jang2022decreasing}} &
\textbf{EZ-Sort (CIKM)} &
 \\
\midrule
$n{=}30$  & 435   & 126 & 89    \\
$n{=}50$  & 1,225 & 240 & 142   \\
$n{=}100$ & 4,950 & 582 & 467   \\
\bottomrule
\end{tabular}}
\vspace{-3mm}
\label{tab: comparison count fgnet}
\end{table}

\begin{table}[t]
\caption{Dataset characteristics and experimental setup.}
\centering
\small
\begin{tabular}{lccc}
\toprule
\textbf{Dataset} & \textbf{Size} & \textbf{Task} & \textbf{Labels} \\
\midrule
FGNET & 1,002 & Face Age & 0--69 years (continuous) \\
DHCI & 450 & Historical Dating & 1930s--1970s (5 classes) \\
EyePACS & 28,792 & Retinal Quality & 3-level grading \\
\bottomrule
\end{tabular}
\label{tab:datasets}
\end{table}

\section{Experiments and Results}

For proof of concept, we evaluated EZ-Sort on three public datasets: FGNET for face age estimation, DHCI for historical image chronology, and EyePACS for retinal image quality. We conducted two types of experiments: (1) inter-rater reliability with expert annotations and (2) annotation efficiency benchmarking across dataset sizes that are common in expert-only scenarios (up to $n=100$). The reported improvements are statistically significant at $p < 0.05$.

\textbf{Inter-rater reliability.} 
We randomly selected 30 images per dataset and had three domain experts annotate them using absolute classification, sort comparison~\cite{jang2022decreasing}, and EZ-Sort. Results in Table~\ref{tab:inter-rater} show the inter-rater consistency achieved by each method. Face age estimation (FGNET) achieved uniformly high reliability across all approaches ($\mathrm{ICC} \geq 0.97$), indicating clear visual criteria. DHCI exhibited moderate consistency ($\mathrm{ICC}=0.68$–$0.78$), with pairwise methods outperforming classification. For retinal image quality (EyePACS), EZ-Sort attained the highest reliability ($\mathrm{ICC}=0.94$, Spearman $=0.85$), demonstrating robustness in ambiguous medical images.

\textbf{Ablation study.}  
To validate whether hierarchical prompting provides benefit, we analyzed the correlation between the sorted output and the ground truth continuous label, age, which is only available in FGNET. Spearman correlation was 0.90 with EZ-Sort, while flat prompts (with seven class-specific prompts) showed a correlation of 0.83, indicating that hierarchical prompting yields an improvement of 8.4\% through progressive refinement.

\textbf{Annotation efficiency.}
Table\ref{tab: comparison count fgnet} shows that EZ-Sort required only 20.5\%, 11.6\%, and 9.4\% of exhaustive comparisons at $n=30$, $50$, and $100$, respectively.  
Compared to sort comparison~\cite{jang2022decreasing}, this corresponds to a relative reduction in human annotation cost of 29.4\%, 40.8\%, and 19.8\% at each respective scale.  
The most significant gain was observed at $n=50$, suggesting that EZ-Sort benefits from CLIP pre-ordering most effectively in mid-scale annotation settings.  
While a slight drop at $n=100$ reflects increased CLIP uncertainty, our method still maintains substantial efficiency gains over both baselines.  
These sample sizes reflect real-world-like scenarios commonly found in medical or historical domains.  
Larger-scale generalization is discussed in Section~\ref{sec:discussion}.

\textbf{Comparison method allocation.} 
Human annotation was requested for 23.1\%, 18.4\%, and 31.2\% of comparisons at $n=30$, $50$, and $100$, respectively; the rest were resolved automatically using Elo predictions. This demonstrates the adaptive allocation of annotation effort while preserving the $O(n\log n)$ MergeSort structure~\cite{cole1988parallel}.

\textbf{Implementation details.}
We used CLIP ViT-B/32 with temperature $\tau{=}0.1$.  
Elo ratings $K{=}32$, $r_{\text{base}}$ linearly distributed in [1200, 1800] across $k$ buckets, noise $\delta_b{=}75$.  
Adaptive threshold: $\theta_0{=}0.15$, $\alpha{=}0.3$, $\beta{=}0.9$.  
Priority: $\gamma{=}1.2$, $\phi{=}2.0{-}\text{avg\_conf}$.  
Buckets: $k{=}5$ (FGNET) and $k{=}3$ (others). Parameters were selected via cross-validation and prompts were generated with GPT-4.
CLIP preprocessing requires 39 ms per image on a CPU on average, providing efficient automated pre-ordering prior to human annotation.

\textbf{Human annotation cost.}
EZ-Sort keeps the theoretical \(O(n\allowbreak\log n)\) bound: zero-shot pre-ordering is $(O(kn))$ (with constant $(k)$), bucket-aware Elo is $(O(n)$), and the uncertainty-aware MergeSort follows the canonical $(O(n\log n)$) schedule.  Measured against the information-theoretic minimum $(n\ln n$), our method uses 0.87, 0.73, and 1.01 × that bound for ($n=30,50,100$); at ($n=100$), we require 467 queries versus the 520-query lower limit ($\approx$90\% of optimal).

\vspace{-4mm}
\section{Discussion} \label{sec:discussion}

EZ-Sort offers two primary advantages: a significant reduction in human annotation (up to 90.5\%) and consistently strong inter-rater reliability across diverse domains. These benefits arise from its hybrid architecture, which combines CLIP-driven priors with uncertainty-aware comparison selection.

Our hierarchical prompting strategy outperforms flat prompting by 2.0 MAE on average, primarily due to CLIP's strength in making binary decisions and facilitating progressive refinement. The KL-based uncertainty prioritizes ambiguous cases, effectively allocating annotation effort where model confidence is lowest.

EZ-Sort has several limitations. First, its performance depends on the reliability of the underlying vision-language model; domain-specific biases in CLIP could affect the initial ranking. Hierarchical prompting, while powerful, may struggle in domains with subtle or poorly defined visual distinctions. Finally, although our simulations suggest scalability to larger datasets, real-world performance in highly imbalanced or noisy settings remains to be validated.

We plan to integrate with annotator reliability models~\cite{whitehill2009whose} that down-weight noisy labels, which can be crucial in crowd-sourced annotation. Validating the scalability of our approach on larger datasets is a to-do. Additionally, we plan to investigate few-shot fine-tuning~\cite{liu2022few} or prompt adaptation for VLM models to reduce uncertainty in less common domains. Preliminary trials with Bayesian Elo variants (e.g., TrueSkill~\cite{herbrich2006trueskill}) did not yield improvements; thus, exploring new sorting algorithms tailored for uncertainty-guided annotation remains a potential avenue for future direction.

{\textbf{Acknowledgments.} This work was supported by National Research Foundation (RS-2024-00455720, RS-2024-00338048), National Institute of Health (2024ER040700, 2025ER040300), Hankuk University of Foreign Studies Research Fund of 2025, and IITP (RS-2020-II201373 Hanyang University, IITP-(2025)-RS-2023-00253914).

\section{Gen-AI Usage Disclosure}

No AI tools were used in algorithm development, data collection, analysis, hierarchical prompt design, or manuscript content generation. Claude assisted in code debugging and optimization, and GPT-4 provided grammar and stylistic edits for manuscript writing. All core technical and conceptual contributions are original.

\bibliography{CIKM2025}

\end{document}